\documentclass[11pt,a4paper]{article}
\usepackage[hyperref]{emnlp2020}
\usepackage{times}
\usepackage{latexsym}
\usepackage{amsmath}
\usepackage{tabularx,ragged2e}
\usepackage{booktabs}
\usepackage{xurl}
\usepackage{graphicx}
\usepackage{subcaption}
\usepackage{caption}
\usepackage{float}
\usepackage[english]{babel}

\usepackage{multirow} 

\usepackage{todonotes}

\aclfinalcopy %

\title{RobBERT: a Dutch RoBERTa-based Language Model}

\author{Pieter Delobelle$^1$ \and Thomas Winters$^1$ \and Bettina Berendt$^{1,2}$\\
$^1$ Department of Computer Science, KU Leuven \\
$^2$ Faculty of Electrical Engineering and Computer Science, TU Berlin \\
\texttt{firstname.lastname@kuleuven.be}}

\date{}

\begin{document}
\maketitle
\begin{abstract}
Pre-trained language models have been dominating the field of natural language processing in recent years, and have led to significant performance gains for various complex natural language tasks.
One of the most prominent pre-trained language models is BERT, which was released as an English as well as a multilingual version.
Although multilingual BERT performs well on many tasks, recent studies show that BERT models trained on a single language significantly outperform the multilingual version.
Training a Dutch BERT model thus has a lot of potential for a wide range of Dutch NLP tasks.
While previous approaches have used earlier implementations of BERT to train a Dutch version of BERT, we used RoBERTa, a robustly optimized BERT approach, to train a Dutch language model called RobBERT.
We measured its performance on various tasks as well as the importance of the fine-tuning dataset size.
We also evaluated the importance of language-specific tokenizers and the model's fairness.
We found that RobBERT improves state-of-the-art results for various tasks, and especially significantly outperforms other models when dealing with smaller datasets.
These results indicate that it is a powerful pre-trained model for a large variety of Dutch language tasks.
The pre-trained and fine-tuned models are publicly available to support further downstream Dutch NLP applications.
\end{abstract}

\section{Introduction}

The advent of neural networks in natural language processing (NLP) has significantly improved state-of-the-art results within the field.
Initially, recurrent neural networks and long short-term memory networks dominated the field.
Later, the transformer model caused a revolution in NLP by dropping the recurrent part and only keeping attention mechanisms \citep{vaswaniAttention2017}.
The transformer model led to other popular language models, e.g. GPT-2 \citep{radford2018gpt,radford2019gpt2}.
BERT \citep{devlinBERT2019a} improved over previous models and recurrent networks by allowing the system to learn from input text in a bidirectional way, rather than only from left-to-right or the other way around.
This model was later re-implemented, critically evaluated and improved in the RoBERTa model \citep{liuRoBERTa2019}.

These large-scale attention-based models provide the advantage of being able to solve NLP tasks by having a common, expensive pre-training phase, followed by a smaller fine-tuning phase.
The pre-training happens in an unsupervised way by providing large corpora of text in the desired language.
The second phase only needs a relatively small annotated dataset for fine-tuning to outperform previous popular approaches in one of a large number of possible language tasks.

While language models are usually trained on English data, some multilingual models also exist.
These are usually trained on a large quantity of text in different languages.
For example, Multilingual-BERT is trained on a collection of corpora in 104 different languages \citep{devlinBERT2019a}, and generalizes language components well across languages \citep{pires2019multilingual}.
However, models trained on data from one specific language usually improve the performance of multilingual models for this particular language \citep{martinCamemBERT2019,devriesBERTje2019}.
Training a RoBERTa model~\citep{liuRoBERTa2019} on a Dutch dataset thus also potentially increases performances for many downstream Dutch NLP tasks.
In this paper, we introduce RobBERT\footnote{The model named itself RobBERT when it was prompted with \textit{``Ik heet $<$mask$>$BERT.''} (\textit{``My name is $<$mask$>$BERT.''}), which we found quite a suitable name.}, a Dutch RoBERTa-based pre-trained language model, and critically evaluate its performance on various language tasks against other Dutch languages models.
We also propose several new tasks for testing the model's zeroshot ability, evaluate its performance on smaller datasets, and for measuring the importance of a language-specific tokenizer.
Finally, we provide an extensive fairness evaluation using recent techniques and a new translated dataset.

\section{Related Work}
\label{sec:related-work}

Transformer models have been successfully used for a wide range of language tasks.
Initially, transformers were introduced for use in machine translation, where they efficiently improved the state-of-the-art \citep{vaswaniAttention2017}.
This cornerstone was used in BERT, a transformer model obtaining state-of-the-art results for eleven natural language processing tasks, such as question answering and natural language inference \citep{devlinBERT2019a}.
BERT is pre-trained with large corpora of text using two unsupervised tasks.
The first task is called masked language modeling (MLM),
making the model guess which word is masked in certain position in the text.
The second task is next sentence prediction, in which the model has to predict if two sentences occur subsequent in the corpus, or randomly sampled from the corpus.
These tasks allow the model to create internal representations about a language, which could thereafter be reused for different language tasks.
This architecture has been shown to be a general language model that could be fine-tuned with little data in a relatively efficient way for a very distinct range of tasks and still outperform previous architectures \citep{devlinBERT2019a}.

Transformer models are also capable of generating contextualized word embeddings \citep{peters2018elmo}.
Traditional word embeddings, e.g. word2vec~\citep{mikolovEfficient2013} and GloVe~\citep{penningtonGlove2014}, lack the capibility of differentiating words based on context (e.g. \emph{``a stick''} versus \emph{``let's stick to''}).
Transformer models, like BERT, on the other hand automatically incorporate the context a word occurs into its embedding.

The attention mechanism in transformer encoder models also allows for better resolution of coreferences between words \citep{joshi2019spanbert}.
For example, in the sentence
\emph{``The trophy doesn’t fit in the suitcase because it’s
too big.''}, the word \emph{``it''} would refer to the the suitcase instead of the trophy if the last word was changed to \emph{``small''} \citep{levesque2012winograd}.
Being able to resolve these coreferences is for example important for translation, as dependent words might change form, e.g. due to word gender.

While BERT has been shown to be a powerful language model, it also received scrutiny on its training and pre-processing.
The authors of RoBERTa \citep{liuRoBERTa2019} showed that while the NSP pre-training task made the model perform better, it was not due to its intended reason, as it might merely predict relatedness between corpus sentences rather than subsequent sentences.
That \citet{devlinBERT2019a} trained a better model when using NSP than without NSP is likely due to the model learning long-range dependencies that were longer than when just using single sentences.
As such, the RoBERTa model uses only the MLM task, and uses multiple full sentences in every input.
Other researchers later improved the NSP task by instead making the model predict for two subsequent sentences if they occur in the given or flipped order in the corpus \citep{lan2019albert}.

\citet{devlinBERT2019a} also presented a multilingual model (mBERT) with the same architecture as BERT, but trained on Wikipedia corpora in 104 languages.
Unfortunately, the quality of these multilingual embeddings is considered worse than their monolingual counterparts, as
\citet{ronnqvistMultilingual2019} illustrated for German and English models in a generative setting.
The monolingual French CamemBERT model~\citep{martinCamemBERT2019} also outperformed mBERT on all tasks. 
\citet{brandsen2019bert} also outperformed mBERT on several Dutch tasks using their Dutch BERT-based language model, called BERT-NL, trained on the small SoNaR corpus \citep{oostdijk2013sonar}.
More recently, \citet{devriesBERTje2019} also showed similar results for Dutch using their BERTje model, outperforming multilingual BERT in a wide range of tasks, such as sentiment analysis and part-of-speech tagging by pre-training on multiple corpora.
Since both these works are concurrent with ours, we compare our results with BERTje and BERT-NL in this paper.

\section{Pre-training RobBERT}

We pre-trained RobBERT using the RoBERTa training regime.
We trained two different versions, one where only the pre-training corpus was replaced with a Dutch corpus \emph{(RobBERT v1)} and one where both the corpus and the tokenizer were replaced with Dutch versions \emph{(RobBERT v2)}.
These two versions allow to evaluate the importance of having a language-specific tokenizer.

\subsection{Data}

We pre-trained our model on the Dutch section of the OSCAR corpus, a large multilingual corpus which was obtained by language classification in the Common Crawl corpus \citep{ortizsuarezAsynchronous2019}.
This Dutch corpus is 39GB large, with 6.6 billion words spread over 126 million lines of text, where each line could contain multiple sentences.
This corpus is thus much larger than the corpora used for similar Dutch BERT models, as BERTje used a 12GB corpus, and BERT-NL used the SoNaR-500 corpus (about 2.2GB). \citep{devriesBERTje2019,brandsen2019bert}.

\subsection{Tokenizer}

For RobBERT v2, we changed the default byte pair encoding (BPE) tokenizer of RoBERTa to a Dutch tokenizer.
The vocabulary of the Dutch tokenizer was constructed using the Dutch section of the OSCAR corpus~\citep{ortizsuarezAsynchronous2019} with the same byte-level BPE algorithm as RoBERTa~\citep{liuRoBERTa2019}. %
This tokenizer gradually builds its vocabulary by replacing the most common consecutive tokens with a new, merged token.
We limited the vocabulary to 40k words, which is 10k words less than RobBERT v1, due to additional tokens including non-negligible number of Unicode tokens that are not used in Dutch.
These are likely caused due to misclassified sentences during the creation of the OSCAR corpus~\citep{ortizsuarezAsynchronous2019}.

\subsection{Training}

RobBERT shares its architecture with RoBERTa's base model, which itself is a replication and improvement over BERT \citep{liuRoBERTa2019}.
Like BERT, it's architecture consists of 12 self-attention layers with 12 heads \citep{devlinBERT2019a} with 117M trainable parameters.
One difference with the original BERT model is due to the different pre-training task specified by RoBERTa, using only the MLM task and not the NSP task.
During pre-training, it thus only predicts which words are masked in certain positions of given sentences.
The training process uses the Adam optimizer~\citep{kingmaAdam2017} with polynomial decay of the learning rate $l_r=10^{-6}$ and a ramp-up period of 1000 iterations, with hyperparameters $\beta_1=0.9$ %
and RoBERTa's default $\beta_2=0.98$.
Additionally, a weight decay of 0.1 and a small dropout of 0.1 helps prevent the model from overfitting \citep{srivastavaDropout2014}.

RobBERT was trained on a computing cluster with 4 Nvidia P100 GPUs per node, where the number of nodes was dynamically adjusted while keeping a fixed batch size of 8192 sentences.
At most 20 nodes were used (i.e. 80 GPUs), and the median was 5 nodes. %
By using gradient accumulation, the batch size could be set independently of the number of GPUs available, in order to maximally utilize the cluster. %
Using the Fairseq library~\citep{ott2019fairseq}, the model trained for two epochs, which equals over 16k batches in total, which took about three days on the computing cluster.
In between training jobs on the computing cluster, 2 Nvidia 1080 Ti's also covered some parameter updates for RobBERT v2.

\section{Evaluation}

We evaluated RobBERT on multiple downstream Dutch language tasks.
For testing text classification, we evaluate on sentiment analysis and on demonstrative and relative pronoun prediction.
The latter task helps evaluating the zero-shot prediction abilities, i.e. using only the pre-trained model without any fine-tuning.
Both classification tasks are also used to measure how well RobBERT performs on smaller datasets, by only using subsets of the data.
For testing RobBERT's token tagging capabilities, we used both part-of-speech (POS) tagging and named entity recognition (NER) tasks.

\begin{table*}[tbh]
\centering
\caption{Results of RobBERT fine-tuned on several downstream classification tasks, compared to the state of the art models for the tasks. For accuracy, we also report the 95\% confidence intervals. \textit{(Results annotated with * from \citet{vanderburghMerits2019}, ** from \citet{devriesBERTje2019}, *** from \citet{brandsen2019bert}, **** from \citet{alleinBinary2020})}}
\label{tab:results}
\resizebox{\textwidth}{!}{%
\begin{tabular}{@{}lll|ll@{}}
\toprule
                                                 & \multicolumn{2}{c}{\textbf{10k}}       & \multicolumn{2}{c}{\textbf{Full dataset}}       \\ 
Task + model                                             & ACC (95\% CI) [\%]   & F1 [\%] & ACC (95\% CI) [\%]    & F1 [\%] \\
\midrule
\textbf{Sentiment Analysis (DBRD)}                      &                               &                &                                &         \\
\multicolumn{1}{r}{\citet{vanderburghMerits2019}}       & ---                           & ---            & 93.8*                          & ---         \\
\multicolumn{1}{r}{BERTje~\citep{devriesBERTje2019}}    & ---                           & ---            & 93.0**                         & ---         \\
\multicolumn{1}{r}{BERT-NL~\citep{brandsen2019bert}}    & ---                           & ---            & ---                            & 84.0***        \\
\multicolumn{1}{r}{RobBERT v1}                      &  86.730  (85.32, 88.14)       & 86.729         & 94.422 (93.47,95.38)  & 94.422       \\
\multicolumn{1}{r}{RobBERT v2}                      &  \textbf{94.379}  (93.42, 95.33)       & \textbf{94.378} & \textbf{95.144} (94.25,96.04)  & \textbf{95.144}       \\

\textbf{Die/Dat (Europarl)}                             &                               &                &                                &         \\
\multicolumn{1}{r}{Baseline~\citep{alleinBinary2020}}    & ---                           & ---            & 75.03****                       & ---       \\
\multicolumn{1}{r}{mBERT~\citep{devlinBERT2019a}}        & 92.157 (92.06, 92.25)          & 90.898         & 98.285 (98.24,98.33)           & 98.033                        \\
\multicolumn{1}{r}{BERTje~\citep{devriesBERTje2019}}    & 93.096 (92.84, 93.36)         & 91.279         & 98.268 (98.22,98.31)         & 98.014                         \\
\multicolumn{1}{r}{RobBERT v1}                      & 97.006 (96.95, 97.07)& 96.571 & {98.406} (98.36, 98.45) & {98.169} \\
\multicolumn{1}{r}{RobBERT v2}                      & \textbf{97.816} (97.76, 97.87)& \textbf{97.514}& \textbf{99.232} (99.20, 99.26) & \textbf{99.121} \\ \bottomrule
\end{tabular}%
} 
\end{table*}

\subsection{Sentiment Analysis}

We replicated the high-level sentiment analysis task used to evaluate BERT-NL~\citep{brandsen2019bert} and BERTje \citep{devriesBERTje2019} to be able to compare our methods.
This task uses a dataset called Dutch Book Reviews dataset (DBRD), in which book reviews from \url{hebban.nl} are labeled as positive or negative \citep{vanderburghMerits2019}.
Although the dataset contains 118,516 reviews, only 22,252 of these reviews are actually labeled as positive or negative, which are split in a 90\% train and 10\% test datasets.
This dataset was released in a paper analysing the performance of an ULMFiT model (Universal Language Model Fine-tuning for Text Classification model) \citep{vanderburghMerits2019}.

We fine-tuned RobBERT on the first 10,000 training examples as well as on the full dataset.
While the ULMFiT model is first fine-tuned using the unlabeled reviews before training the classifier \citep{vanderburghMerits2019}, it is unclear whether the other BERT models utilized the unlabeled reviews for further pre-training \citep{sun2019classification} or only used the labeled data for fine-tuning the pre-trained model.
We did the latter, meaning further improvement is possible by additionally pre-training on unlabeled in-domain sequences.
Another unknown is how these models dealt with reviews that were longer than the maximum number of tokens, 
as the average book review length is 547 tokens, with 40\% of the documents being longer than our model could handle.
For our experiments, we only gave the last tokens of a review as input, as we found the training performance to be better,
likely due to containing a summarizing comments.
We trained our model for 2000 iterations with a batch size of 128 and a warm-up of 500 iterations, reaching a learning rate of $10^{-5}$. 
The training took approx. 2 hours on 2 Nvidea 1080 Ti GPUs, the best-performing RobBERT v2 model was selected based on a validation accuracy of 0.994. 
We see that RobBERT outperforms the other BERT models.
Both versions of RobBERT also outperform the state-of-the-art ULMFiT model, although the difference is only statistically significant for RobBERT v2.

\subsection{Die/Dat Disambiguation}\label{ss:die-dat}

Since BERT models perform well on coreference resolution tasks \citep{joshi2019coreference}, we propose to evaluate RobBERT on the recently introduced \emph{``die/dat disambiguation''} task \citep{alleinBinary2020}, as a novel way to evaluate the zeroshot ability of Dutch BERT models.
In Dutch, depending on the sentence, both \emph{``die''} and \emph{``dat''} can be either demonstrative or relative pronouns; in addition they can also be used in a subordinating conjunction, i.e. to introduce a clause.
The use of either of these words depends on the gender of the word it refers to.
\citet{alleinBinary2020} presented multiple models trained on the Europarl~\citep{koehnEuroparl2005a} and SoNaR corpora~\citep{oostdijkConstruction2013}, achieving an accuracy of 75.03\% on Europarl to 84.56\% on SoNaR.

For this task, we use the Dutch Europarl corpus~\citep{koehnEuroparl2005a}, with the first 1.3M sequences ({\tt head}) for training and last 399k ({\tt tail}) as test set. 
Every sequence containing \emph{``die''} or \emph{``dat''} creates an example for every occurrence of either word by masking the occurrence.
For the test set, this resulted in about 289k masked sentences.

BERT-like models can solve this task using two different approaches.
Since the task is about predicting words, their default MLM task can be used to guess which of the two words is more probable in a particular masked position.
This allows the comparison of zero-shot BERT models, i.e. without any fine-tuning on the training data (\autoref{tab:zeroshot}).
The second approach uses the masked sentences to create two versions by filling the mask with either ``die'' and ``dat'', separate them using the \texttt{[SEP]} token and making the model predict which of the two sentences is correct.
This fine-tuning was performed using 4 Nvidia GTX 1080 Ti GPUs, %
taking 30 minutes for 13 epochs on 10k sequences and about 24 hours for 3 epochs on the full dataset.
We did no hyperparameter tuning, as the initial hyperparameters ($l_r=10^{-5}, \epsilon=10^{-9}$, warm-up of 250 steps, batch size of 32 (10k) or 128 (full dataset), dropout of 0.1) were satisfactory. 

To measure RobBERTs performance on smaller datasets, we trained the model twice for both the sentiment analysis task and the \emph{die/dat} disambiguation task, once with a subset of 10k utterances, and once with the full training dataset.

\begin{table}[htb]
\centering
\caption{Performance of predicting \textit{die/dat} as most likely candidate for a mask using zero-shot BERT models (i.e. without fine-tuning)  as well as a majority class predictor (ZeroR), tested on the 288,799 test set sentences}
\label{tab:zeroshot}
\resizebox{0.9\linewidth}{!}{%
\begin{tabular}{@{}lr@{}}
\toprule 
\textbf{Model}                                       & \textbf{Accuracy [\%]}  \\
\midrule
\multicolumn{1}{r}{ZeroR (majority class)}   & 66.70       \\
\multicolumn{1}{r}{mBERT~\citep{devlinBERT2019a}}    & 90.21  \\
\multicolumn{1}{r}{BERTje~\citep{devriesBERTje2019}}& 94.94     \\
\multicolumn{1}{r}{RobBERT v1}                  & 98.03  \\ 
\multicolumn{1}{r}{RobBERT v2}                  & \textbf{98.75}  \\ 
\bottomrule
\end{tabular}%
} 
\end{table}

RobBERT outperforms previous models as well as other BERT models both with as well as without fine-tuning (see Table \ref{tab:results} and Table \ref{tab:zeroshot}).
It is also able to reach similar performance using less data.
The fact that both for the fine-tuned and the zero-shot setting, RobBERT outperforms other BERT models is also an indication that the base model has internalised more knowledge about Dutch than the others, likely due to the improved pre-training regime and using a larger corpus.
We can also see that having a Dutch tokenizer strongly helps reduce the error rate for this task, halving the error rate when fine-tuned on the full dataset.
The reason the BERT-based models outperform the previous RNN-based approach is likely the encoders ability to better deal with coreference resolution \citep{joshi2019spanbert}, and by extension deciding which word the \textit{``die''} or \textit{``dat''} belongs to.
The fact that RobBERT strongly outperforms the other BERT models on subsets of the data indicates that it is a suitable candidate for Dutch tasks that only have limited data available.

\subsection{Part-of-speech Tagging}
\label{ss:pos}

Part-of-speech (POS) tagging involves labeling tokens rather than labeling sequences.
For this, we used a different head with an classification output for each token, all activated by a softmax function. 
When a word consists of multiple tokens, the first token is used for the the label of the word.

We perform the same POS fine-tuning regimes as RoBERTa \citep{liuRoBERTa2019} to evaluate RobBERT's performance.
When fine-tuning, we employ a linearly decaying learning rate with a warm-up for 6\% of the total optimisation steps \citep{liuRoBERTa2019}.
For all the encoder-based models in our evaluation, we also perform a limited hyperparameter search on the development set with learning rate $l_r \in \{10^{-5}, 2\cdot 10^{-5}, 3\cdot 10^{e-5}, 10^{-4}\}$ and batch size~$\in \{16, 32, 48\}$, which is also based on RoBERTa's fine-tuning.

\begin{table}[tbh]
\centering
\caption{POS tagging on Lassy UD. For accuracy, we also report the 95\% confidence intervals.}
\label{tab:results-tokens}

\resizebox{\linewidth}{!}{%
\begin{tabular}{@{}lll@{}}
\toprule
Task + model                                         & ACC (95\% CI) [\%]           \\ \midrule
\multicolumn{1}{r}{Frog~\citep{bosch2007frog}}     & 91.7 (91.2, 92.2)  \\

\multicolumn{1}{r}{mBERT~\citep{devlinBERT2019a}}     & \textbf{96.5} (96.2, 96.9)   \\
\multicolumn{1}{r}{BERTje~\citep{devriesBERTje2019}} & 96.3 (96.0, 96.7)            \\
\multicolumn{1}{r}{RobBERT v1}                   & 96.4 (96.0, 96.7)            \\
\multicolumn{1}{r}{RobBERT v2}                   & 96.4 (96.0, 96.7)            \\ \bottomrule
\end{tabular}
}
\end{table}

To evaluate the POS-performance, we used the Universal Dependencies (UD) version of the Lassy dataset \citep{vanNoord2013lassy}, containing 17 different POS tags.
We compared its performance with Frog, a popular memory-based Dutch POS tagging approach, and with other BERT models.
Surprisingly, multilingual BERT marginally outperformed both Dutch BERT models, although not statistically significantly, with both RobBERT models in second place with an almost equal accuracy.
The higher performance of multilingual BERT could be indicative that it benefits from transferable language structures from other languages helping it to perform well for POS tagging.
Alternatively, this could signal a limit of the UD Lassy dataset, or at least for the performance of BERT-like models on this dataset.

\begin{figure}
    \centering
    \includegraphics[width=\linewidth]{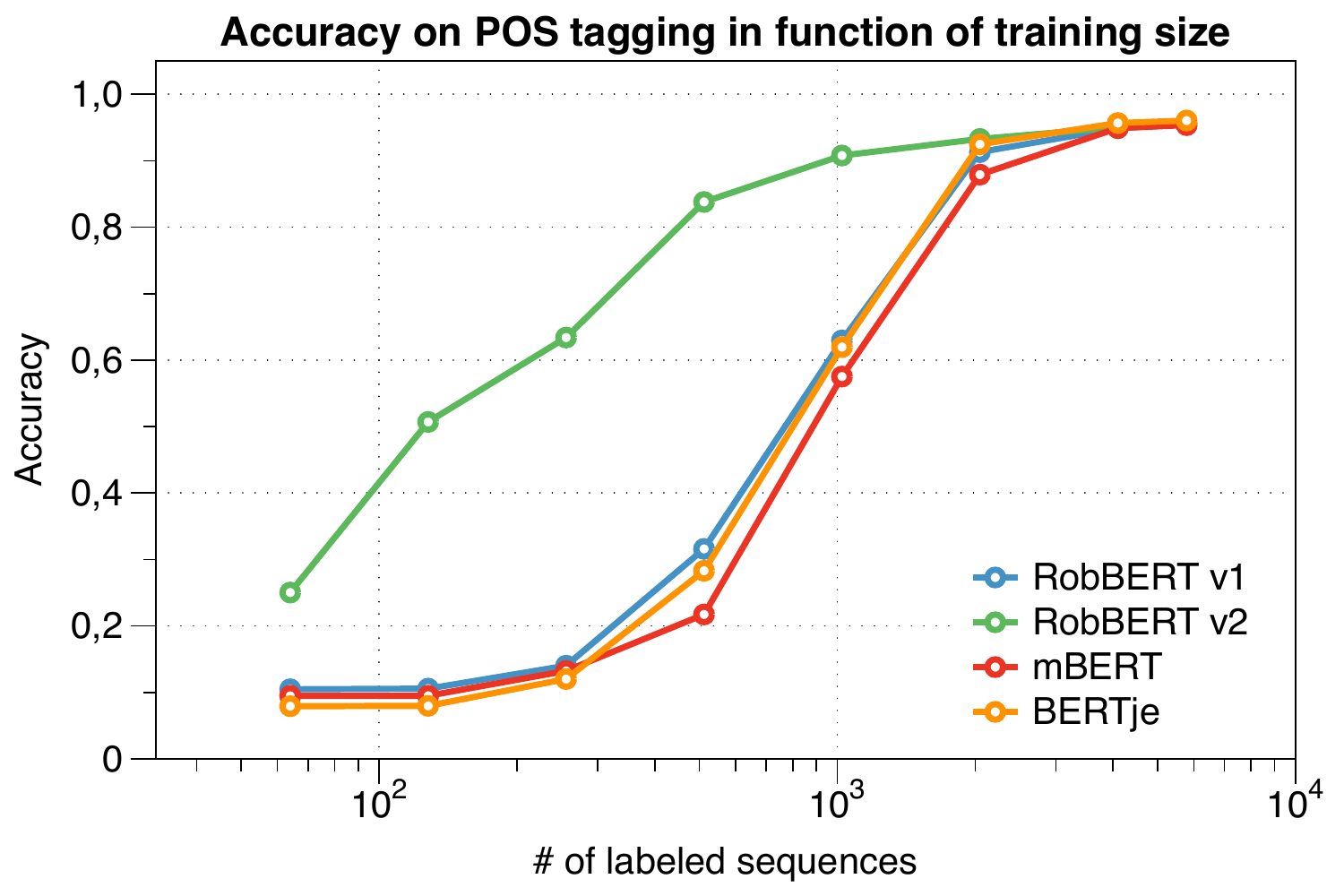}
    \caption{POS tagging accuracy on the test set for different sizes of training sets.}
    \label{fig:robbert-pos-acc}
\end{figure}

We also evaluated the models on several smaller subsets of the training data, to illustrate how much data is needed to achieve acceptable results.
For all models, the same hyperparameters obtained for \autoref{tab:results-tokens} are used for all subsets, under the assumption that using a subset of the training data also works well under the same hyperparameters.
The hyperparameters which yielded the results of RobBERT v2 are $l_r=10^{-4}$, batch size of 16 and dropout of 0.1.
The separate development set was used to select the best-performing model after each epoch based , which had a cross-entropy loss of 0.172 on the development set.
While all BERT models perform similarly after seeing all instances of the UD Lassy dataset, there is a clear difference when using smaller training sets (\autoref{fig:robbert-pos-acc}).
RobBERT v2 outperforms all other models when using only 1,000 data points or less, again showing that it is more capable of dealing with smaller datasets.

\subsection{Named Entity Recognition}

Named entity recognition (NER) is the task of labeling named entities in a sentence.
It is thus a token-level task, just like POS-tagging, meaning we can use the same setup and hyperparameter tuning as described in \autoref{ss:pos}.
We use the CoNLL-2002 dataset and evaluation script\footnote{Retrieved from \url{https://www.clips.uantwerpen.be/conll2002/ner/}}, which use a four value BIO labeling, namely for organisations, locations, people and miscellaneous \citep{sang2002conll}.
The hyperparameters yielding the results for RobBERT v2 are $l_r=3\cdot10^{-5}$, batch size of 32 and dropout of 0.1.
The separate development set was used to select the best-performing model after each epoch.
As the $F_1$ score is generally used for evaluation of this task, we opted to use this metric instead of cross-entropy loss for selecting the best-performing model, which had an $F_1$ score of 0.8769 on the development set.
We compared the $F_1$ scores on the NER task in \autoref{tab:results-tokens-ner}.

\begin{table}[htb]
\centering
\caption{NER for various models, $F_1$ score calculated with the CoNLL 2002 evaluation script, except for $\dagger$ which used the Seqeval Python library, * from \citet{wuBeto2019}, ** from \citet{brandsen2019bert}, *** from \citet{devriesBERTje2019}.}
\label{tab:results-tokens-ner} 

\resizebox{0.9\linewidth}{!}{%
\begin{tabular}{@{}lll@{}}
\toprule
Task + model                                         & $F_1$ score [\%]     \\ \midrule
\multicolumn{1}{r}{Frog~\citep{bosch2007frog}}     & 57.31   \\
\multicolumn{1}{r}{mBERT~\citep{devlinBERT2019a}}     & 84.19  \\
\multicolumn{1}{r}{mBERT \citep{wuBeto2019}}     & \textbf{90.94}*   \\
\multicolumn{1}{r}{BERT-NL~\citep{brandsen2019bert}}     & 89.7$^\dagger$**  \\
\multicolumn{1}{r}{BERTje~\citep{devriesBERTje2019}} & 88.3***  \\
\multicolumn{1}{r}{RobBERT v1}                   & 87.53 \\
\multicolumn{1}{r}{RobBERT v2}                   & 89.08 \\ \bottomrule
\end{tabular}
}
\end{table}

We can see that \citep{wuBeto2019} outperforms all other BERT models using a multilingual BERT model with an $F_1$ score of 90.94.
When we used the token labeling fine-tuning regime described earlier on multilingual BERT, we were only able to get to an $F_1$ score of 84.19 using multilingual BERT, thus being outperformed by the Dutch BERT models.
One possibility is that the authors used a more optimal fine-tuning regime, or that they trained their model longer.

\section{RobBERT and Fairness}

As language models are trained on large corpora, this poses a risk that minorities and protected groups are ill-represented, e.g. by encoding stereotypes~\citep{bolukbasi2016man, zhaoGender2019, gonen-goldberg-2019-lipstick}.
In word embeddings, these studies often rely on analogies~\citep{bolukbasi2016man,caliskanSemantics2017}
or embedding analysis~\cite{gonen-goldberg-2019-lipstick}.
These approaches are not directly transferable to BERT models, 
since the sentence the word occurs in influences its embedding.

Efforts to generalize these approaches often rely on templates~\citep{mayMeasuring2019, kurita-etal-2019-measuring}.
These can be intentionally neutral (``\emph{{\tt \textless mask\textgreater} is a word}'') or they might resemble an analogy in textual form (``\emph{{\tt \textless mask\textgreater} is a zookeeper.}'').
One can then perform an association test between possible values for the {\tt \textless mask\textgreater} slot, similar to a word embedding association test~\citep{caliskanSemantics2017}.

In this section, we discuss two distinct potential origins of representational harm \citep{blodgettLanguage2020} a language model could exhibit, and evaluate these on RobBERT v2.
The two discussed behaviours are (i) stereotyping of gender roles in occupations %
and (ii) unequal predictive power for texts written by men and women. %
These exemplifications highlight how language models risk affecting the experience of the end user, or replicating and reinforcing stereotypes. 

\subsection{Gender Stereotyping}
\label{ss:stereotyping-bias}

To assess how gender stereotypes of professions are present, we performed a template-based association test similar to \citet{kurita-etal-2019-measuring} and the \emph{semantically unbleached} templates of \citet{mayMeasuring2019}. 
We used RobBERT's LM head---trained during pre-training with the MLM task---to fill in the \emph{\tt \textless mask\textgreater} slot for each template, in the same manner as the zero-shot task described in \autoref{ss:die-dat}.
These templates have a second slot, which is used to iterate over the professions. 

For this list of professions and the gender projection on the \emph{he-she} axis, we base us on the work by \citet{bolukbasi2016man}, who crowdsourced the associated gender for various professions. 
Ideally, we would use a similarly crowdsourced Dutch dataset. However, since this does not yet exist, we opted for manually translating these English professions using the guidelines established by the European Parliament for gender neutral professions~\citep{dimitriospapadimoulisGenderneutraal2018},
meaning that we opted for the inclusive form for neutral professions in English that do not have a neutral counterpart, but an inclusive binary male variant and a female variant with explicit gender (e.g. for lawyer: using \emph{``advocaat''} and not \emph{``advocate''}).
In the rare case that an inclusive or neutral form translated to an exclusive binary form, we excluded this profession. 

\begin{figure}[tb]
    \makebox[1.05\linewidth][r]{ %
    \centering
    \includegraphics[width=1.125\linewidth]{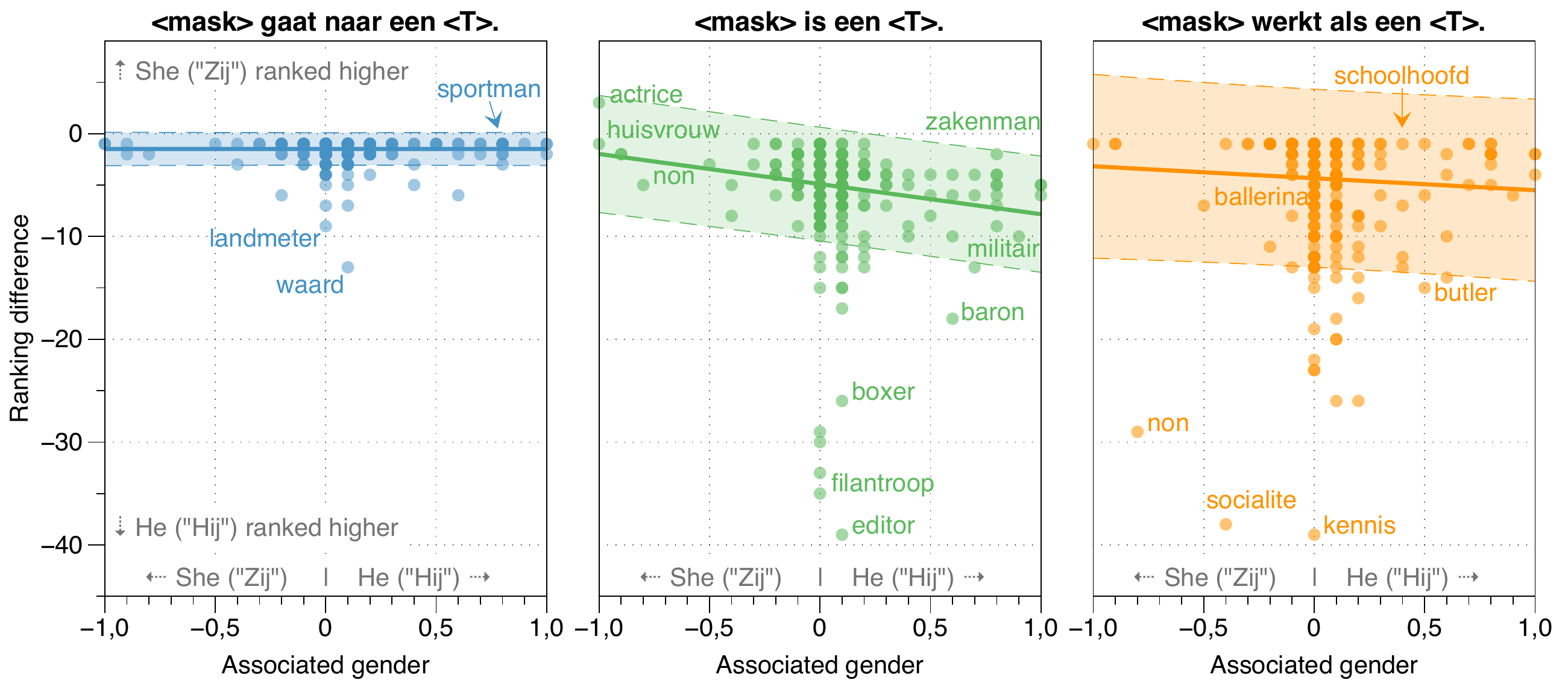}
    }
    \caption{Ranking difference between gendered pronouns for various professions. Three templates were used to evaluate, were {\tt \textless T\textgreater} is replaced by a profession. In the leftmost template, the pronoun and profession refer to different entities.} 
    \label{fig:gender-diff-lm}
\end{figure}

We evaluated three templates on RobBERT, including one control template without co-referent entities (``{\tt \textless mask\textgreater} \emph{goes to a} {\tt \textless T\textgreater}'') (\autoref{fig:gender-diff-lm}).
For the control template, there should be no and indeed is no  correlation between ranking difference for both pronouns and the associated gender of a profession.
Interestingly, none of the instances has a positive ranking difference, meaning the language model always ranks the male pronoun as more likely.

When the profession and {\tt \textless mask\textgreater} slot refer to the same entity, the general assessment of the language model correlates with the associated gender.
But again, RobBERT estimates that the male pronoun is more likely in almost all cases, even when these professions have a gendered suffix.
Curiously, actress (``\emph{actrice}'') is the only word where this is not the case.
Since RobBERT estimates the male pronoun to be more likely even in the control template, we suspect that the effect is due to more coverage of men in the training corpus.

\subsection{Unequal Predictive Performance}
\label{ss:system-performance-bias}

Unfairness is particularly problematic if it leads to unequal predictive performance. This problem has been demonstrated for decision support systems, including recidivism prediction~\citep{angwinMachine2016} and public employment services~\citep{allhutterAlgorithmic2020}. Such predictions can be downstream tasks of language understanding; for example when job resumés are processed \citep{ van-hautte-etal-2020-leveraging}. 

To review fairness in downstream tasks, we evaluated the sentiment analysis task on DBRD, a dataset with scraped book reviews. 
Although this task in itself may have low impact for end users, it still serves as an illustrative example of how fine-tuned models can behave unfairly.

To investigate whether such bias might result for our fine-tuned model, we analyzed its outcome for different values of a sensitive attribute (in this case gender), as is commonly done in fair machine learning research~\citep{zemel2013learning, hardtEqualityOpportunitySupervised2016, delobelleEthical2020}.
To this end, we augmented the held-out test set of DBRD with gender as a sensitive attribute for each review\footnote{We make this augmentation of DBRD available under CC-by-NC-SA at \url{https://people.cs.kuleuven.be/~pieter.delobelle/data.html}.}.
Values were obtained from the reviews' author profiles with a self-reported binary gender (\emph{`man'} or \emph{`vrouw'}) (64\%). The remaining 36\% of reviews did not report 
author
gender, and they were discarded for this evaluation.
Of the remaining, gender-labelled, reviews, 76\% were written by women.
Thus, the dataset is unbalanced.

We quantify the gender difference with two metrics: (i) Demographic Parity Ratio (DPR), which expresses a relative difference between predicted outcomes $\hat{y}$ conditioned on the sensitive attribute $a$~\citep{dworkFairnessAwareness2012}, following

\[
\frac{P( \hat{y} \mid \neg a)}{P( \hat{y} \mid a)},
\]

and (ii) Equal Opportunity (EO) %
\citet{hardtEqualityOpportunitySupervised2016}, 
which in addition also conditions on the true outcome $y$,
as a task-specific %
fairness
measure \citep{dworkFairnessAwareness2012}, following 
\[
P( \hat{y} \mid \neg a, y) - P( \hat{y} \mid a, y).
\]

\begin{figure}[t]
    \centering
    \includegraphics[width=1.05\linewidth]{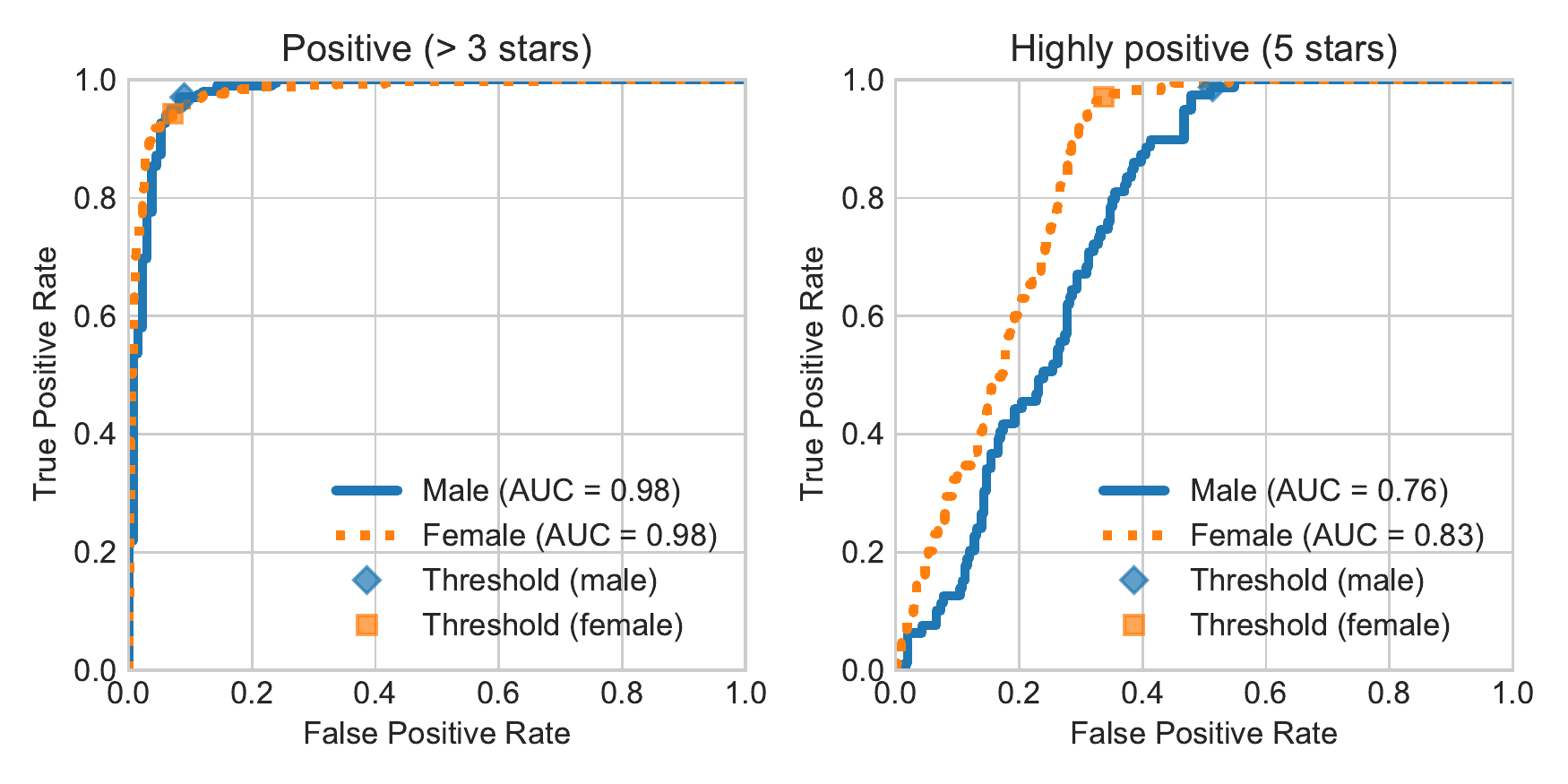}
    \caption{ROC of the fine-tuned model to predict positive reviews for male and female reviewers}
    \label{fig:dbrd-gender}
\end{figure}

\citet{hardtEqualityOpportunitySupervised2016} also relate EO to the ROC curves to evaluate fairness when dealing with a binary predictor and a score function.
To derive a binary predictor, 
we used 0 as a threshold value. \autoref{fig:dbrd-gender} shows the single resulting predictor, with the ROC curves split on the sensitive attribute, for each of the two rating levels (over 3 resp. 5 stars).

The results of \autoref{fig:dbrd-gender} show that there is small difference in opportunity,
which is especially pronounced for the highly positive classifier.
For positive reviews, the EO difference is 0.028 at the indicated threshold and DPR is 70.2\%.
The DPR would indicate an 
unfairness, as values below 80\% are often considered unfair.
However, this metric has received some 
criticism,
and when including the true outcome in EO, the difference in probabilities is close to 0,
which does not signal any unfairness.
When taking into account the ROC curves (\autoref{fig:dbrd-gender}), the EO score can be explained by the good predictive performance. 
When considering highly positive reviews, however, the differences become more pronounced and the model has better predictive performance for reviews written by women. 

\section{Code}

The training and evaluation code of this paper as well as the RobBERT model and the fine-tuned models are publicly available for download at \url{https://github.com/iPieter/RobBERT}.

\section{Limitations and Future Work}

There are several potential improvements for creating a better pre-trained RobBERT-like model.
First, since BERT-based models are still 
being %
actively researched, one could potentially improve the training regime using new unsupervised pre-training tasks when they are discovered, e.g. sentence order prediction \citep{lan2019albert}.
Second, while RobBERT is trained on lines that contain multiple sentences, it does not put subsequent lines of the corpus after each other due to the shuffled nature of the OSCAR corpus \citep{ortizsuarezAsynchronous2019}.
This is unlike RoBERTa, which does put full sentences next to 
one another
if they do not exceed the available sequence length, in order to learn the long-range dependencies between words that the original BERT learned using its controversial NSP task.
Creating an unshuffled version of OSCAR might thus further improve the performance of the pre-trained model.
Third, there might be some benefit to modifying the tokenizer to use morpheme-based tokens, as Dutch uses compound words.
Fourth, one could improve model's fairness during pre-training.
We illustrated how representational harm in downstream tasks can affect the end user's experience, like the unequal predictive performance for the DBRD task.
Various methods have been proposed to mitigate unfair behaviour in AI models~\citep{zemel2013learning, delobelleEthical2020}.
While we refrained from training fair pre-trained and fine-tuned models in this research, training such models could be an interesting contribution.
In addition, with the increased attention on fairness in machine learning, a broader view of the impact on other protected groups due to large pre-trained language models is also called-for.

The RobBERT model itself can be used in new settings to help future research.
First, RobBERT could be used in a model that uses a BERT-like transformer stack for the encoder and a generative model as a decoder~\citep{raffelExploring2019,lewis2019bart}
Second, RobBERT can serve as the basis for a large number of Dutch language tasks that we did not examine in this paper.
Given RobBERT's state-of-the-art performance %
on small as well as on large datasets,
it could help advance results
when fine-tuned on new datasets. %

\section{Conclusion}

We introduced a new language model for Dutch based on RoBERTa, called RobBERT, and showed that it outperforms earlier approaches as well as other BERT-based language models for a several different Dutch language tasks.
More specifically, we found that RobBERT significantly outperformed other BERT-like models when dealing with smaller datasets, making it a useful resource for a large range of application domains.
We expect this model to serve as a base for fine-tuning on other tasks, and thus help foster new models that can advance results for Dutch language tasks.

\section*{Acknowledgements}

Pieter Delobelle was supported by the Research Foundation - Flanders under EOS No. 30992574 and received funding from the Flemish Government under the “Onderzoeksprogramma Artificiële Intelligentie (AI) Vlaanderen” programme.
Thomas Winters is a fellow of the Research Foundation-Flanders (FWO-Vlaanderen).
Most computational resources and services used in this work were provided by the VSC (Flemish Supercomputer Center), funded by the Research Foundation - Flanders (FWO) and the Flemish Government – department EWI.
We are especially grateful to Luc De Raedt for his guidance as well as for providing the facilities to complete this project.
We are thankful to Liesbeth Allein and her supervisors for inspiring us to use the \textit{die/dat} task.
We are also grateful to \citet{ott2019fairseq, paszke2019pytorch, Haghighi2018, wolfHuggingFace2019} for their software packages.

\bibliography{main,main2} 
\bibliographystyle{acl_natbib}

\end{document}